\documentclass[runningheads]{llncs}
\usepackage{graphicx}
\usepackage{tikz}
\usepackage{comment} 
\usepackage{amsmath,amssymb} 
\usepackage{color}
\usepackage{multirow}       
\usepackage{url}
\usepackage{wrapfig}

\begin{document}
\pagestyle{headings}
\mainmatter
\def\ECCVSubNumber{4257}  

\title{Describing Unseen Videos via Multi-Modal Cooperative Dialog Agents} 


\titlerunning{Describing Unseen Videos via Multi-Modal Cooperative Dialog Agents}
%
\author{Ye Zhu\inst{1} \and
Yu Wu\inst{2,3} \and
Yi Yang\inst{2} \and
Yan Yan\inst{1}\thanks{Corresponding author.}
}
\authorrunning{Y. Zhu et al.}
%

\institute{Texas State University, San Marcos, USA \\
\email{\{ye.zhu, tom\_yan\}@txstate.edu}\\
\and
ReLER, University of Technology Sydney, Australia \\
\email{yu.wu-3@student.uts.edu.au,yi.yang@uts.edu.au}
\and 
Baidu Research, China
}
\maketitle

\begin{abstract}
With the arising concerns for the AI systems provided with direct access to abundant sensitive information, researchers seek to develop more reliable AI with implicit information sources. To this end, in this paper, we introduce a new task called video description via two multi-modal cooperative dialog agents, whose ultimate goal is for one conversational agent to describe an unseen video based on the dialog and two static frames. Specifically, one of the intelligent agents - \emph{Q-BOT} - is given two static frames from the beginning and the end of the video, as well as a finite number of opportunities to ask relevant natural language questions before describing the unseen video. \emph{A-BOT}, the other agent who has already seen the entire video, assists \emph{Q-BOT} to accomplish the goal by providing answers to those questions. We propose a QA-Cooperative Network with a dynamic dialog history update learning mechanism to transfer knowledge from \emph{A-BOT} to \emph{Q-BOT}, thus helping \emph{Q-BOT} to better describe the video.
Extensive experiments demonstrate that \emph{Q-BOT} can effectively learn to describe an unseen video by the proposed model and the cooperative learning method, achieving the promising performance where \emph{Q-BOT} is given the full ground truth history dialog. Codes and models are available at \url{https://github.com/L-YeZhu/Video-Description-via-Dialog-Agents-ECCV2020}.

\keywords{Video Description, Dialog Agents, Multi-Modal.}
\end{abstract}

\section{Introduction}
\label{sec:intro}

It is becoming a trend to exploit the possibilities to develop artificial intelligence (AI) systems with subtle and advanced reasoning abilities as humans, usually by providing AI with direct access to abundant information sources. However, what comes with this tendency is the arising concerns over the security and privacy issues behind such AI systems. Although direct access to rich information assists AI becoming more intelligent and competent, the general public starts to question whether their sensitive personal information, such as the identifiable face images and voices, is in safe hands.

Despite the general concerns, nowadays research in AI and computer vision (CV) fields is experiencing a rapid transition from traditional `low-level' tasks within single modality data such as image classification~\cite{krizhevsky2012imagenet,he2016deep,he2015delving}, object detection~\cite{redmon2016you,szegedy2013deep,lin2017feature}, machine translation~\cite{bahdanau2015neural}, to more challenging tasks that involve multiple modalities of data and subtle reasoning~\cite{gan2019multi,yang2016stacked,hudson2018compositional,song2018explore,santoro2017simple}, such as visual question answering~\cite{anderson2018bottom,antol2015vqa,xu2016ask} and visual dialog~\cite{das2017visual,jain2018two,wu2018you}. 
A meaningful and informative conversation, either between human-computer or computer-computer, is an appropriate task to demonstrate such a reasoning process due to the complex information exchange mechanism during the dialog. With the emergence of large-scale datasets such as VQA~\cite{antol2015vqa}, GuessWhat~\cite{de2017guesswhat} and AVSD~\cite{alamri2019audio,hori2019end}, much effort is devoted to study the techniques for machines to maintain natural conversation interactions in a sophisticated way~\cite{jain2018two,wu2018you,lee2018answerer,massiceti2018flipdial}. While most existing works still focus on the dialog itself and only involve a single agent, generally by providing the agent with direct access to sensitive information (\textit{e.g.}, the complete video clips with identifiable human faces and voices), we wish to take a step forward to more secure and reliable AI systems with implicit information sources.
To this end, in this paper we introduce a novel natural and challenging task with implicit information sources: describe an unseen video mainly based on the dialog between two cooperative agents.

\begin{figure}[t]
    \centering
    \includegraphics[width=0.75\textwidth]{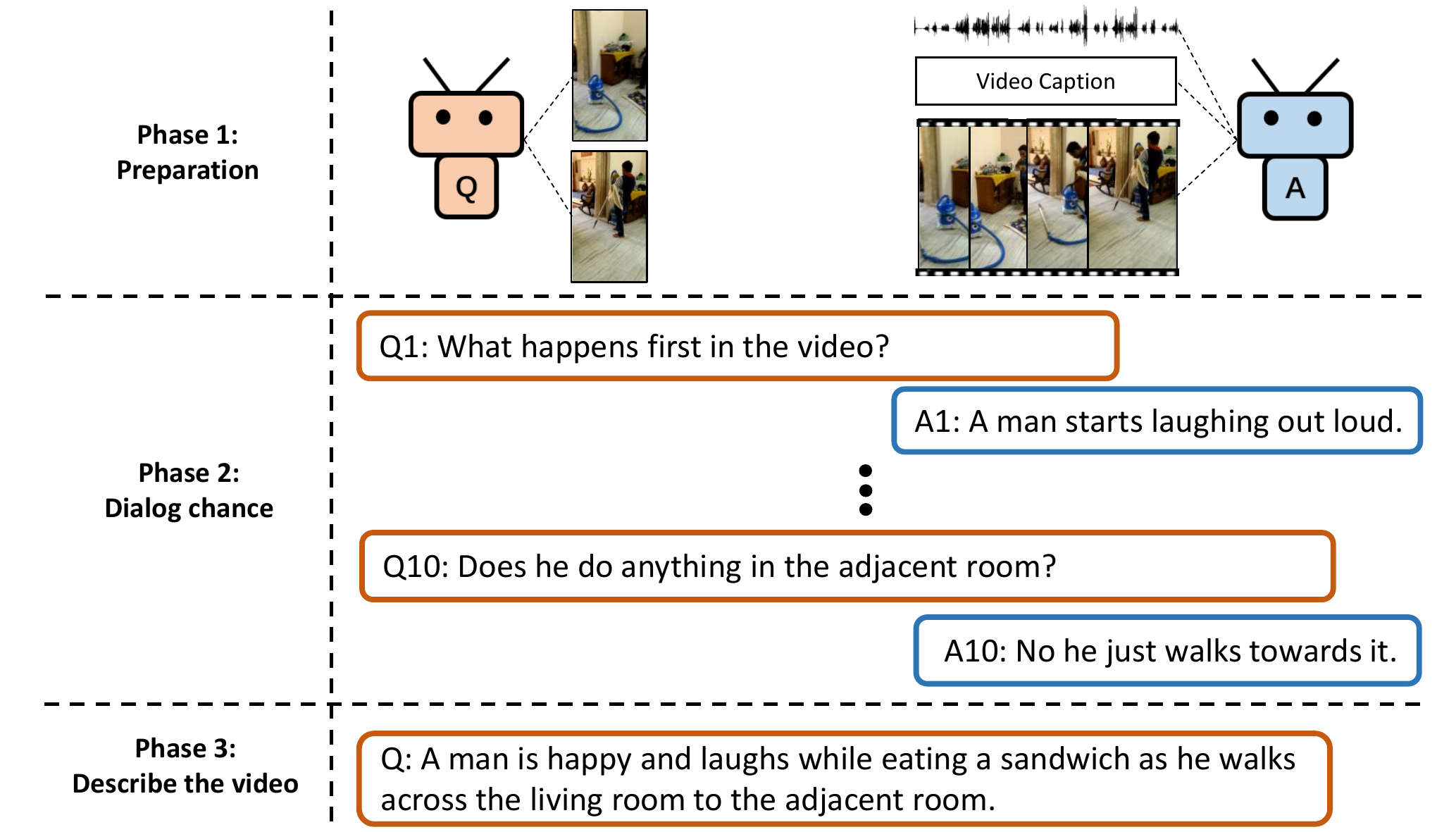}
    \caption{Task setup: Describing an unseen video via two multi-modal cooperative dialog agents. The entire process can be described in three phases. The ultimate goal is for \emph{Q-BOT} to describe what happens between the beginning and the end of an unseen video based on the dialog history with \emph{A-BOT}.}
    \label{fig:1}
\end{figure}

\begin{figure}[t]
    \centering
    \includegraphics[width=0.98\textwidth]{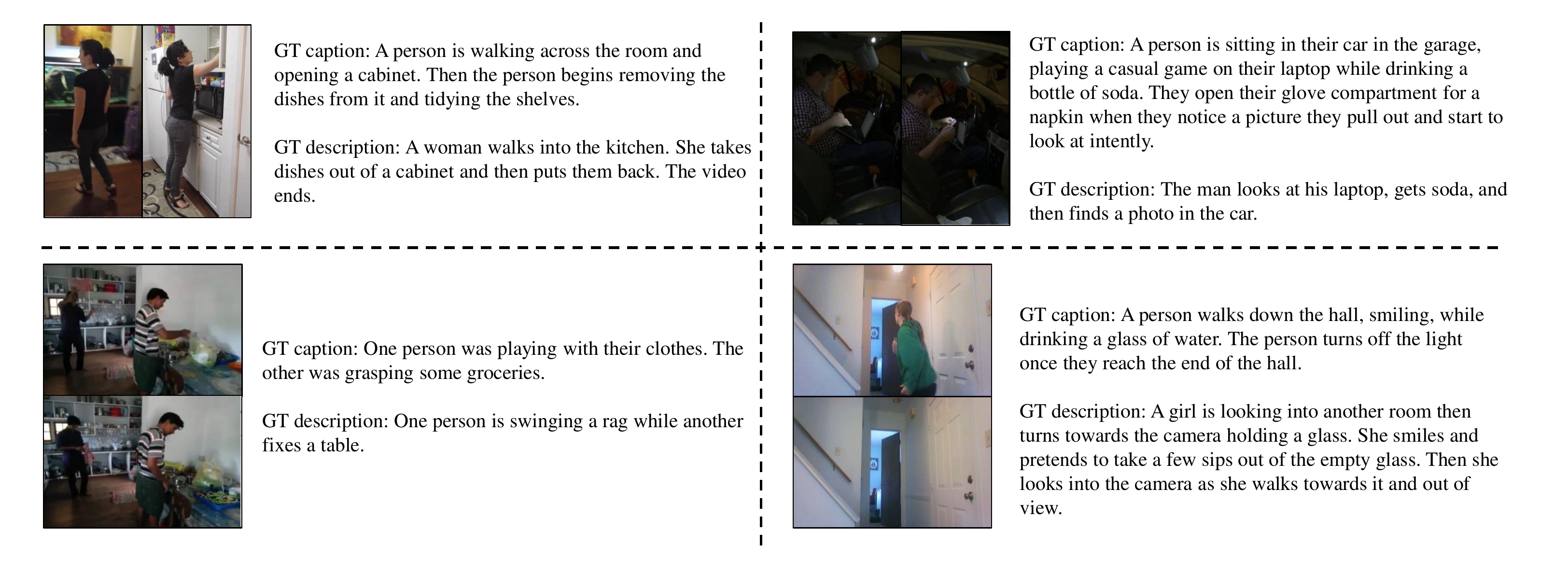}
    \caption{Examples from the AVSD dataset~\cite{alamri2019audio,hori2019end}. Two static images are the beginning and the end frame of the video clips. \textit{GT caption} is the caption shown to the \emph{A-BOT} under our task setup. \textit{GT description} is actual the summary given by the human questioner at the end of the dialog during the data collection without directly watching the video, which corresponds to the video description required in our task. 
    Intuitively, our task can be considered as establishing an additional information barrier(\textit{i.e.}, the dialog) between the direct visual data input and the natural language caption output.}
    \label{fig:2}
\end{figure}

The setup for our task is illustrated in Fig.~\ref{fig:1}, which involves two dialog agents, \emph{Q-BOT} and \emph{A-BOT}. Imagine the \emph{Q-BOT} to be the actual AI system, and the \emph{A-BOT} to be humans.
The entire process can be described in three phases: 
In the first preparation phase, two agents are provided with different information. \emph{A-BOT} is able to see the complete video with the audio signals and captions, while \emph{Q-BOT} is only given two static frames from the beginning and the end of the video. 
It is worth noting that the static frames given to \emph{Q-BOT} have no specific requirements, they could be static images without visible humans or just the back of the person as shown in Fig.~\ref{fig:1} and Fig.~\ref{fig:2}, which largely reduces the risks for the AI systems to recognize the actual person.
In the second phase, \emph{Q-BOT} has 10 opportunities to ask \emph{A-BOT} relevant questions about the video, such as the event happened. After 10 rounds of question-answer interactions, \emph{Q-BOT} is asked to describe the unseen video based on the initial two frames and the dialog history with \emph{A-BOT}. Under this task setup, the AI system \emph{Q-BOT} accomplishes a multi-modal task without direct access to the original information, but learns to filter and extract useful information from a less sensitive information source, \textit{i.e.}, the dialog. It is highly impossible for AI systems to identify a person based on the natural language descriptions. Therefore, such task settings and reasoning ability based on implicit information sources have great potential to be applied in a wide practical context, such as the smart home systems, improving the current AI systems that rely on direct access to sensitive information to accomplish certain tasks. Notably, instead of directly asking for the final video descriptions from human users, our task formulation that requires AI systems to gradually ask questions helps to reduce the bias and noises usually contained in the description directly given by human individuals.

It is a challenging and natural task compared to previous works in the field of the visual dialog. The difficulty mainly comes from the implicit information source of multiple modalities and the more complex reasoning process required for both agents. 
Specifically, this task also differs from the traditional video captioning task due to the fact that \emph{Q-BOT} has never seen the entire video. Intuitively, it can be considered as establishing an additional information barrier(\textit{i.e.}, the dialog) between the direct visual data input and the natural language video caption output.
Fig.~\ref{fig:2} shows some examples from the AVSD dataset~\cite{alamri2019audio,hori2019end}, whose data collection process resembles to our task setup. We observe that the ground truth video captions shown to \emph{A-BOT} and the expected final descriptions given by \emph{Q-BOT} are quite different, thus revealing the actual gap existing in human reasoning. This fact again emphasizes the difficulties of our task.

The key aspect to consider in this work is the effective knowledge transfer from \emph{A-BOT} to \emph{Q-BOT}. \emph{A-BOT}, who plays the role of humans, has full access to all the information, while \emph{Q-BOT} only has an ambiguous understanding of the surrounding environment from two static video frames after the first phase. In order to describe the video with details that are not included in the initial input, \emph{Q-BOT} needs to extract useful information from the dialog with \emph{A-BOT}. Therefore, we propose a QA-Cooperative network that involves two agents with the ability to process multiple modalities of data. We further introduce a cooperative learning method that enables us to jointly train the network with a dynamic dialog history update mechanism. The knowledge gap and transfer process are both experimentally demonstrated.

The main contributions of this paper are: 
1) We propose a novel and challenging video description task via two multi-modal dialog agents, whose ultimate goal is for one agent to describe an unseen video mainly based on the interactive dialog history. This task establishes a more secure and reliable setting by providing implicit information sources to AI systems. 2) We propose a QA-Cooperative network and a cooperative learning method with a dynamic dialog history update mechanism, which helps to effectively transfer knowledge between the two agents. 3) We experimentally demonstrate the knowledge gap as well as the transfer process between two agents on the AVSD dataset~\cite{hori2019end}. With the proposed method, our \emph{Q-BOT} achieves very promising performance comparable to the strong baseline situation where full ground truth dialog is provided.

\section{Related Work}

\noindent \textbf{Image and Video Captioning.} Image or video captioning refers to the task that aims to obtain textual descriptions for the given image or video. It is one of the first well-exploited tasks that combines both computer vision and natural language processing. 
You \textit{et al.}~\cite{you2016image} adopt Recurrent neural networks (RNNs) with selective attention to semantic concept proposals to generate image captions. 
Adaptive attention and spatial attention have also been studied in~\cite{lu2017knowing,chen2017sca}.
Anderson \textit{et al.}~\cite{anderson2018bottom} propose to use bottom-up and top-down attention for object levels and other salient image regions to address this task.
Yao \textit{et al.}~\cite{yao2017boosting} use the attributes to further improve the performance. 
Rennie \textit{et al.}~\cite{rennie2017self} propose to train the image captioning system with reinforcement learning.
Wang \textit{et al.}~\cite{wang2017diverse} adopt conditional variational auto-encoders with an additive Gaussian encoding space to generate image descriptions.
Wu~\textit{et al.}~\cite{wu2018decoupled} propose a disentangled framework to generalize image captioning models to describe unseen objects.
RecNet~\cite{wang2018reconstruction} is introduced for video captioning. Mahasseni\textit{et al.}~\cite{mahasseni2017unsupervised} proposes to summarize the video with adversarial lstm networks in an unsupervised manner.

\noindent \textbf{Visual Question Answering.} Another similar research field beyond describing an image or a video is visual question answering (VQA). The objective of VQA is to answer a natural language question about the given image~\cite{antol2015vqa}.  
Different attention mechanisms including hierarchical attention~\cite{lu2016hierarchical}, question-guided spatial attention~\cite{xu2016ask}, stacked attention~\cite{yang2016stacked}, bottom-up and top-down attention~\cite{anderson2018bottom} have been exploited. Das \textit{et al.}~\cite{das2017human} look into the question whether the existing attention mechanisms attend to the same regions as humans do. Shih \textit{et al.}~\cite{shih2016look} map textual queries and visual features into a shared space to answer the question. Dynamic memory networks is also used for VQA~\cite{xiong2016dynamic}.

\noindent \textbf{Visual Dialog.} Visual dialog is another succession of vision-language problem after visual captioning and VQA that is closely related to our work. Different from the VQA that only involves a single round of natural language interaction, visual dialog requires machines to maintain multiple rounds of conversation. Several datasets have been collected for this task~\cite{das2017visual,de2017guesswhat}. Jain \textit{et al.}~\cite{jain2018two} proposes a symmetric baseline to demonstrate how visual dialog can be generated from discriminative question generation and answering. Additionally, attention mechanisms~\cite{niu2019recursive,seo2017visual} and reinforcement learning~\cite{wu2018you,das2017learning} have also been exploited in the context of visual dialog.

\noindent \textbf{Audio Data.} As another important source of information, the audio modality has recently gained popularity in the research field of computer vision. There have been emerging studies on combining audio and visual information for applications such as sound source separation~\cite{gao2018learning,owens2018audio}, sound source localization~\cite{zhao2018sound,arandjelovic2018objects} and audio-visual event localization~\cite{tian2018audio,wu2019dual}.

\noindent \textbf{Audio-Visual Scene-Aware Dialog.} Audio-visual scene-aware dialog is a recently proposed multi-modal task that combines the previously mentioned research fields. The AVSD dataset introduced in~\cite{hori2019end} and~\cite{alamri2019audio} contains videos with audio streams and a corresponding sequence of question-answer pairs. Hori \textit{et al.}~\cite{hori2019end} enhance the quality of generated dialog about videos using multi-modal features. Schwartz \textit{et al.}~\cite{schwartz2019simple} adopt the multi-modal attention mechanism to extract useful features for audio-visual scene-aware dialog task.

\noindent \textbf{Cooperative Agents and Reasoning.} Das \textit{et al.}~\cite{das2017learning} are the first to propose goal-driven training for dialog agents, during which two interactive dialog agents are also involved to select an unseen image. Wu \textit{et al.}~\cite{wu2018you} propose to generate reasoned dialog via adversarial learning. An information theoretic algorithm for goal-oriented dialog is then introduced in~\cite{lee2018answerer} to help the question generation. Unlike the previous work that has separate agents for questioner and answerer, Massiceti \textit{et al.}~\cite{massiceti2018flipdial} propose to use a single generative model for both roles. The reasoning process is essential in developing such intelligent agents. The idea of multi-step reasoning in the field of VQA has been exploited in~\cite{wu2018you,gan2019multi,yang2016stacked,hudson2018compositional,song2018explore}.

Our work is related to the works mentioned above, yet differentiates from them in multiple aspects, including the task setup and formulation with implicit information sources, the QA-Cooperative network design, and the cooperative training method.

\section{Video Description via Cooperative Agents}

In this section, we respectively introduce the QA-Cooperative network for the two dialog agents, each component of the proposed network and the cooperative learning method with a dynamic dialog history update mechanism.

\begin{table}[t]
\begin{center}
\caption{Notations for the video description task.}
\scalebox{0.85}{
\begin{tabular}{|l|l|}
\hline
$s$ - Video description           &  $V_s$ - start static frame of the video\\ \hline
$\mathcal{S}$ - Vocabulary         &  $V_e$ - end static frame of the video\\ \hline
$i (i \leq 10 )$ - Question-Answer round      &  $x_{A,i}$ - input for \emph{A-BOT} at round $i$\\ \hline
$A$ - Audio data                &$x_{Q,i}$ - input for \emph{Q-BOT} at round $i$  \\ \hline
$V_A$ - Video data for A-BOT &  $r_m$ - original data embedding for modal $m$\\ \hline
$C$ - Video caption     &  $a_m$ - attended data embedding for modal $m$ \\ \hline
$H_{i-1}$ - Existing dialog history at round $i$            &  $d_m$ - dimension of the embedding for modal $m$ \\ \hline
$p_i$ - $i$-th pair of question-answer &  $n_{\{C,H,q,a\}}$ - length of textual sequence\\ \hline
$q_i$ - $i$-th question      &  \multirow{2}{*}{ \begin{tabular}[c]{@{}l@{}}$m$ - modality notation, specified in context\\ $m \in \{A,V,C,H,q,a\}$\end{tabular}} \\ \cline{1-1}
$a_i$ - $i$-th answer  &    \\ \hline 

\end{tabular}}
\end{center}
\label{tab:notation}
\end{table}

Notations used in our task formulation is presented in Table 1.
In this video description task, we expect \emph{Q-BOT} to describe an unseen video with a sentence $s = (s_1, s_2, ..., s_n)$ in $n$ words after 10 rounds of question-answer interactions, each word $s_k$ arises from a vocabulary $\mathcal{S}$. At $i$-th round of question-answer interaction, \emph{A-BOT} takes the audio signals, video data, video caption and the existing dialog history as input, $x_{A,i} = (A, V_A, C, H_{i-1})$, with $H_{i-1} = \{p_1, ..., p_{i-1}\}$ and $p_{i-1} = (q_{i-1}, a_{i-1})$. For \emph{Q-BOT} at the same round $i$, $x_{Q,i} = (V_s, V_e, H_{i-1})$. The final description task for \emph{Q-BOT} is formulated as the inference in a recurrent model with the joint probability given by:

\begin{equation}
    p(s|x_Q) = \prod_{k=1}^{n}p(s_k|s_{<k},x_Q),
\end{equation}
 
 \noindent where we maximize the product of conditionals for each word in description $s$, given the input at 10-th round $x_Q$.

\subsection{QA-Cooperative Network}
\label{subsec:qac}

\begin{figure}[t]
    \centering
    \includegraphics[width=\textwidth]{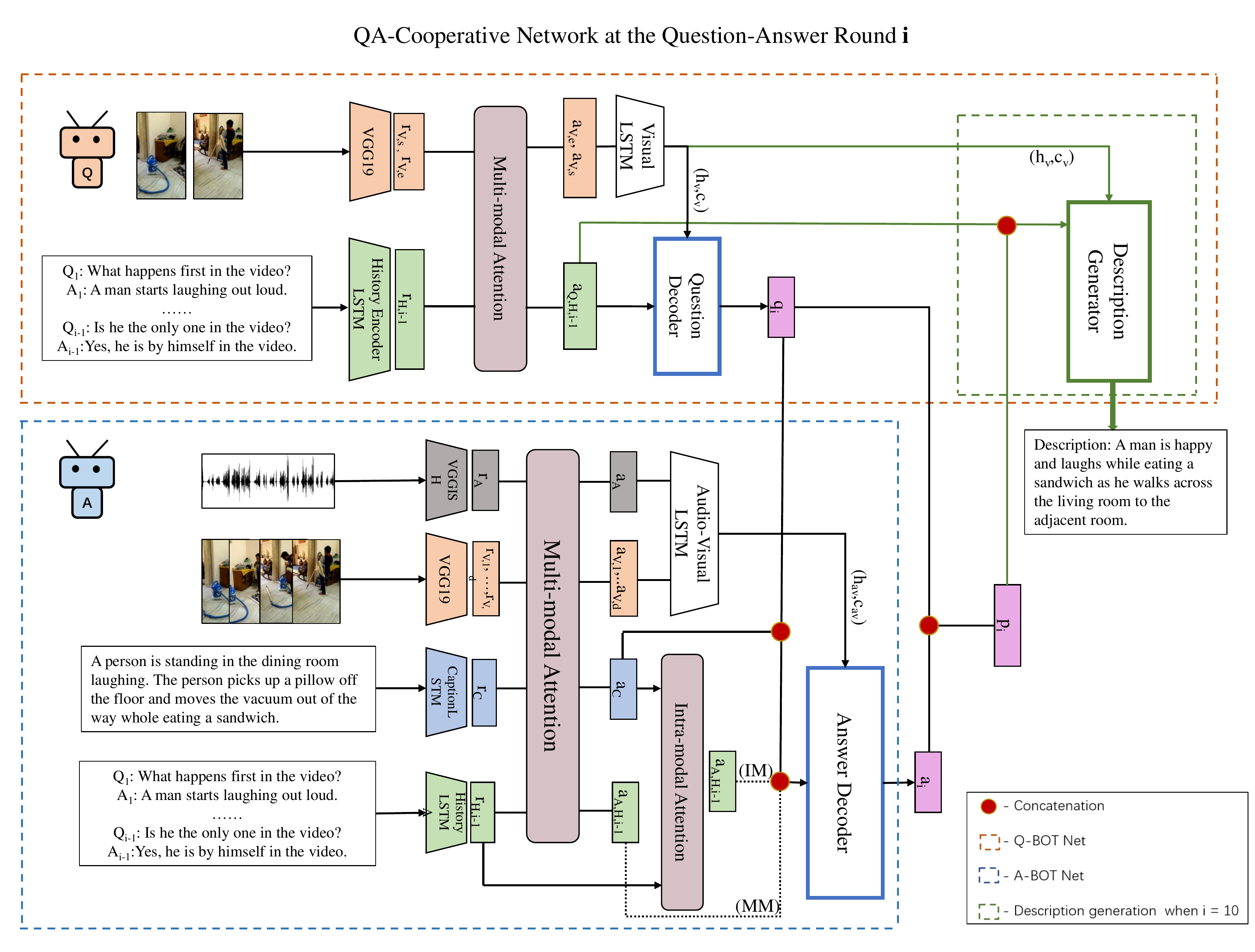}
    \caption{QA-Cooperative Network at the question-answer interaction round $i$. Details about the network architecture and learning method are presented in Sec.~\ref{subsec:qac}, Sec.~\ref{subsec:components} and Sec.~\ref{subsec:learning}.}
    \label{fig:qac}
\end{figure}

The overall architecture of QA-Cooperative network is presented in Fig.~\ref{fig:qac}. \emph{Q-BOT} consists of a visual module, a history encoder, a visual LSTM-net, a multi-modal attention module, a question decoder and the final description generator. \emph{A-BOT} has an audio module, a visual module, a caption encoder, a history encoder, two attention modules and an answer decoder.
In general, \emph{Q-BOT} generates $i$-th question $q_i$ based on the input $x_{Q,i}$, \emph{A-BOT} responds to the question by generating the corresponding answer $a_i$. The new question-answer pair $p_i = (q_i, a_i)$ is used to update the existing dialog history.
Note that we observe from the experiments that separate history decoders for two agents do not help improve the performance. Therefore, we choose the shared history encoder design to reduce the network redundancy. 

\subsection{Model Components}
\label{subsec:components}

There are multiple components in our QA-Cooperative network, which will be explained in detail in this section. We consider the situation at the $i$-th round of question-answer interaction.

\noindent \textbf{Caption Encoder.} The caption encoder contains a linear layer and a single layer LSTM-net. We firstly represent each word in the captions with one-hot vectors. Next, we find the longest caption sentence in each batch and zero-pad other shorter ones. The final caption embedding $r_C \in \mathbb{R}^{n_C \times d_C}$ is obtained from the last hidden state of the LSTM-net. This component is designed for \emph{A-BOT} to encode video captions during the preparation phase. 

\noindent \textbf{History Encoder.} The history encoder contains a linear layer and an LSTM-net. Similarly to the processing steps for the video captions, we start with a list of one-hot word representations for a pair of question-answer. The longest question-answer pair of length is selected, the other pairs are zero-padded to fit the maximum length. The LSTM-net is used to obtain the pair-level embedding $r_{H,i-1} \in \mathbb{R}^{n_T \times n_H \times d_H}$. $n_T$ here denotes the number of question-answer pairs in the dialog history (\textit{i.e., $i-1$}). This encoder is a common component for both \emph{Q-BOT} and \emph{A-BOT}, since the dialog history is visible to both agents. 

\noindent \textbf{Audio-visual LSTM.} It is an LSTM-net with $d+1$ units, where $d$ is the number of visual frames visible to \emph{A-BOT}. It takes the attended audio embedding $a_A \in \mathbb{R}^{d_A}$ and $a_{V,j} \in \mathbb{R}^{d_V}$ with $j = \{1, ...,d\}$ as input, the context vector $(h_{av}, c_{av})$ generated from this LSTM-net is used as the initial states input to the answer decoder. This component is used by \emph{A-BOT} to process the audio and visual information in addition to the cross-modal attention.

\noindent \textbf{Visual LSTM.} Similar to the audio-visual LSTM, this component takes the attended visual embedding $a_{V,s} \in \mathbb{R}^{d_V}$ and $a_{V,e}\in \mathbb{R}^{d_V}$ as input, the context vector $(h_v, c_v)$ from this LSTM-net is used as the initial states for the question decoder and the final description generator. It is used by \emph{Q-BOT} to summarize the visual information from the initial two static frames.

\noindent \textbf{Question Decoder.} The question decoder is formed by an LSTM-net. It takes the attended history embedding $a_{Q,H,i-1} \in \mathbb{R}^{d_H} $ as input, with initial state $(h_0, c_0) = (h_{v}, c_{v})$. The question generator generates the new question $q_i$ that imitates the $i$-th question in the ground truth dialog.

\noindent \textbf{Answer Decoder.} Similar to the question decoder, we use the answer decoder to generate the answer embedding close to the $i$-th answer in the ground truth dialog. The answer LSTM decoder takes the concatenation of the attended history embedding $a_{A,H,i-1} \in \mathbb{R}^{d_H}$, the attended caption embedding $a_c \in \mathbb{R}^{d_C}$ and the newly generated question embedding $r_{q,i}$ as input, with initial state $(h_0, c_0) = (h_{av}, c_{av})$. The output is the answer $a_i$ for the given question. The newly generated question-answer pair at $i$-th round is obtained by combining the $i$-th question and answer. 

\noindent \textbf{Description Generator.} This LSTM generator generates the final description $s$ for the unseen video based on 10 rounds of question-answer interactions history and the two static frames given in the first phase. When $i=10$, the generator computes the following conditional probabilities based on the input, which is the attended history embedding $a_{A,H,10} \in \mathbb{R}^{d_H}$ including 10 rounds of question-answer interactions:

\begin{equation}
    p(s_k|s_{k-1}, h_{k-1}, x_Q) = g(s_k, s_{k-1}, h_{k-1}, x_Q),
\end{equation}

\noindent where $h_{k-1}$ is the hidden states obtained from the previous $k-1$ step. Note that $h$ here is the hidden states of LSTM-net, differnt from the history notation $H$. The initial state is the same as the question decoder, thus we have $(h_0, c_0) = (h_v, c_v)$. the LSTM-net $g$ predicts the probability distribution $p(s_k|s_{k-1}, h_{k-1}, x_Q)$ over words $s_k \in \mathcal{S}_k$, conditioned on the previous words $s_{k-1}$. The final probability distribution for natural language description is obtained by transforming the output of the LSTM-net by a FC-layer and a Softmax.

\noindent \textbf{Attention module.} Since the dialog is a key information source in our task, we propose two different attention mechanisms for processing the information contained in the dialog history: The multi-modal (MM) attention among visual, audio and textual modalities, and the intra-modal (IM) attention between dialog history and another textual sequence. 

For the MM attention, we use the factor graph attention mechanism proposed in~\cite{schwartz2019factor}. For \emph{A-BOT}, this multi-modal attention module takes the audio embedding $r_A$, visual embedding $r_{V,j}$ with $j = \{1,...,d\}$, caption embedding $r_C$ and the history embedding $r_{H,i-1}$ as input. Each visual frame is treated as an individual modality as in~\cite{schwartz2019factor}. The output of this multi-modal attention module are the attended audio embedding $a_A$, the attended visual embedding $a_{V,j}$ with $j = \{1,...,d\}$, and the attended history embedding $a_{Q,H,i-1}$. Similarly for \emph{Q-BOT}, we have the attended output $a_{V,s}$, $a_{V,e}$ and $a_{A,H,i-1}$ after taking $r_{V,s}$, $r_{V,e}$ and $r_{H, i-1}$ as input. Note that the history embedding $r_{H,i-1}$ before the multi-modal attention module is the same for \emph{Q-BOT} and \emph{A-BOT} because a shared history encoder is used, but the attended history embedding becomes different due to different inputs for two agents.

For the IM attention module in Fig.~\ref{fig:qac}, it is a simple softmax attention between the dialog history $r_{H,i-1}$ and the concatenation of $a_{C}$ and $r_{q,i}$.

\subsection{Cooperative Learning}
\label{subsec:learning}

We propose to learn the QA-Cooperative network with a dynamic dialog history update mechanism in a goal-driven manner considering the following two aspects. 

\begin{wrapfigure}[17]{r}{0.5\textwidth}
    \centering
    \includegraphics[width=0.4\textwidth]{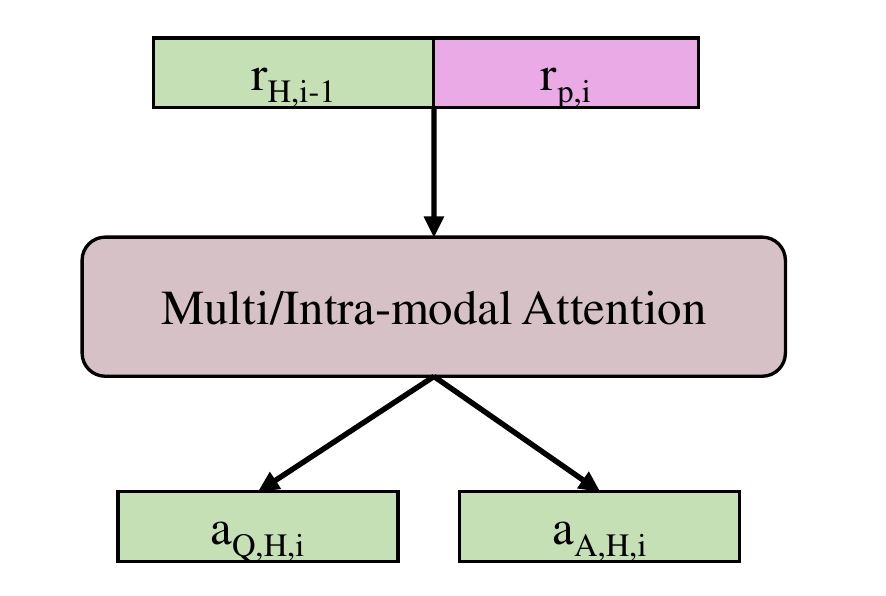}
    \caption{Dialog history update. Specifically, we maintain the history embedding dimension equal to the dimension of $r_{p,i}$ to emphasize the information from the newly generated question-answer pair.}
    \label{fig:fusion}
\end{wrapfigure}

Firstly, the ultimate objective of our task is for \emph{Q-BOT} to describe what happens between the beginning and the end of the video in a concrete way. Considering the fact that only two static frames are given to \emph{Q-BOT} during the first preparation phase, the principal information source for \emph{Q-BOT} is the dialog with \emph{A-BOT}. Therefore, dialog history is the key to effectively learning our QA-Cooperative network. To this end, we propose a dynamic dialog history update mechanism to help with the knowledge transfer from \emph{A-BOT} to \emph{Q-BOT}. Fig.~\ref{fig:fusion} illustrates the dialog history update operation by fusion. Notably, to emphasize the information from the newly generated question-answer pair, we set the dimension of the history embedding equal to the dimension of $r_{p,i}$ before the attention module.

Secondly, for the internal question and answer generation process, we encourage the two agents to imitate the questions and answers from the corresponding rounds of the ground truth dialog. Intuitively, this generation process can also be realized by pre-trained question and answer decoders using the existing methods~\cite{schwartz2019simple,hori2019end,jain2018two,wu2018you} trained in traditional VQA tasks. However, because the pre-training process is directly optimized to generate questions or answers based on the ground truth dialog, it sets the barrier for the final performance in our video description task. In other words, \emph{Q-BOT} is unlikely to surpass the performance obtained in the situation where full ground truth dialog is provided. As for comparisons, the goal-driven training~\cite{das2017learning} is optimized in the final description phase. Although we also use the ground truth dialog as the internal imitation reference for two agents, it leaves more space for flexible question and answer generations (\textit{i.e.}, to ask and answer the questions that directly help \emph{Q-BOT} to better describe the video in the last phase). Our experiments support this assumption in Sec.~\ref{subsec:result}.

Fig.~\ref{fig:qac} schematically illustrates the general learning process at the $i$-th question-answer round. \emph{Q-BOT} takes the two static frames and the existing dialog history that consists of $i-1$ pairs question-answer as input, $x_{Q,i} = (V_s, V_e, H_{i-1})$. The visual and history embedding $r_{V,s}$, $r_{V,e}$ and $r_{H,r-1}$ are processed in the MM attention module~\cite{schwartz2019factor} to acquire the attended embedding $a_{V,s}$, $a_{V,e}$ and $a_{Q,H,r-1}$. Given $(h_v, c_v)$ from the visual LSTM as the initial state, the question decoder outputs the question $q_i$ with its embedding $r_{q,i}$ after taking $a_{Q,H,r-1}$ as input. 
In the meanwhile, \emph{A-BOT} takes the audio signals, video frames, video caption and the existing dialog history as input,  $x_{A,i} = (A, V_A, C, H_{i-1})$. The corresponding attended embedding $a_A$, $a_{V,j}$ with $j={1,...,d}$, $a_C$ and $a_{A,H,i-1}$ is obtained after the MM attention. Another way to obtain the $a_{A,H,i-1}$ is through the IM attention module.
The input for answer decoder is the concatenation of the attended embedding $a_{A,H,i-1}$, $a_C$ and $r_{q,i}$, with initial state $(h_0, c_0)=(h_{av}, c_{av})$ generated from the audio-visual LSTM. The generated question and answer embedding are combined to form the new $i$-th question-answer pair $r_{p,i}$, and used to update the existing history embedding $r_{H,i-1}$ by fusion.

\section{Experiments}

\subsection{Dataset}
\label{subsec: dataset}

We use the recent AVSD v0.1 dataset~\cite{hori2019end} for experiments. The AVSD dataset consists of annotated dialog about 9848 short videos taken from CHARADES~\cite{sigurdsson2016hollywood}. The dialog collection process resembles our task setup, during which two Amazon Mechanical Turk (AMT) workers play the roles of Questioner and Answerer. The Questioner was shown only the first, middle and last static frames of the video, while the Answerer had already watched the entire video, including the audio stream and the original video caption. After having a conversation about the events that happened between the frames through 10 rounds of question-answer interactions, the Questioner is asked to summarize the entire video. 

The current AVSD v0.1 is split into 7659 training, 1787 validation and 1710 testing dialog, respectively. Hori \textit{et al.}~\cite{hori2019end} also propose a `prototype validation-set and test-set', which are sub-splits of the original validation set since the original test set does not include ground truth dialog at the moment. Our experiments are conducted on the original training and `prototype' validation-test splits of the AVSD dataset. 

\subsection{Implementation Details}

\noindent \textbf{Data Representations.} Our cooperative dialog agents have data input of visual, audio and textual modalities. For the visual modal, we take the video representations extracted from the last conv layer of a VGG19 as input. We sample 4 equally spaced frames from the beginning of the original video, and each frame representation is of dimension $7 \times 7 \times 512$. The spatial and visual embedding dimensions are 49 and 512, respectively. \emph{Q-BOT} is shown the first and the last frames, while the \emph{A-BOT} is able to see all the frames. For the audio modal, we obtain the 256-dim audio feature via VGGish~\cite{hershey2017cnn}. For the textual representations, we extract the language embedding from the last hidden state of their corresponding LSTM-nets. The dimensions are $d_C = 256$, $d_q = 128$, $d_a = 128$ and $d_H = 256$.                                

\noindent \textbf{Network Training.} The proposed QA-Cooperative network is trained using a cross-entropy loss on the probabilities $p(s_k|s_{<k}, x_Q)$ on the final video descriptions. All the components are jointly learned. The total amount of trainable parameters is approximately 19M. We use the Adam optimizer with a learning rate of 0.001 and a batch size of 64 for training. During the training, the perplexity metric is used for evaluating the performance on the validation set.

\subsection{Cooperative Video Description}
\label{subsec:result}

We split our experiments into two major groups to provide a more comprehensive and objective analysis of our work as shown in Table 2, which presents the quantitative results using the BLEU1-4~\cite{papineni2002bleu}, METEOR~\cite{banerjee2005meteor}, SPICE~\cite{anderson2016spice}, ROUGE\_L~\cite{lin2004rouge}, and CIDEr~\cite{vedantam2015cider} as the evaluation metrics.

\noindent \textbf{Standard Test Setting.} During our test, the performance of \emph{Q-BOT} is evaluated at each question-answer round-level. In other words, for a given video in the test split, the start question-answer round number $i$ ranges from 1 to 10. For example, if the start round number $i =1$, then no existing dialog history is given to \emph{Q-BOT} and \emph{A-BOT}, they will generate all the ten questions and answers by themselves. However, if the start round number $i=6$, then 5 rounds of question-answer pairs are given to two agents as the existing history, in which case, \emph{Q-BOT} still has 5 opportunities to freely ask questions. For a given video, tests with different start round numbers are independent, representing 10 different test cases. Therefore, for the 733 videos from the `prototype test set' of the AVSD dataset~\cite{hori2019end}, we have in total 7330 different test cases. We refer this testing process as the standard test setting, it is consistent with the learning process explained above in the Sec.~\ref{subsec:learning}.
The only exception is the strong baseline situations for \emph{Q-BOT}, during which the full ground truth dialog history is given before the testing. For the strong baseline situation, since the evaluation is only conducted at the end of full dialog history, there will be 733 testing cases.

\begin{table}[t]
\begin{center}
\caption{Quantitative experimental results of video description tasks by different agents using multiple methods. \textit{HIS Att} stands for \textit{History attention}. The experiments are split into two groups, one group for \emph{A-BOT}, and another group for \emph{Q-BOT}. The comparisons between \emph{A-BOT} and basic \emph{Q-BOT} baselines show the actual knowledge gap between the two agents. 
We obverse that both \emph{A-BOT} and \emph{Q-BOT} from the proposed QA-Cooperative network achieve the best performance. Notably, our \emph{Q-BOT} with cooperative learning is able to achieve comparable performance with the strong baseline, during which the full ground truth dialog history is given as input. It demonstrates the effective knowledge transfer process.}

\scalebox{0.75}{
\begin{tabular}{|c|c|c|c|c|c|c|c|c|c|c|}
\hline
Agent & Method & HIS Att  & BLEU1  & BLEU2    & BLEU3   & BLEU4        & METEOR  & SPICE       & ROUGE\_L      & CIDEr    \\ \hline
\multirow{6}{*}{\begin{tabular}[c]{@{}c@{}}A-BOT\\ (with info from\\  all modalities)\end{tabular}} & Hori \textit{et al.}~\cite{hori2019end}   & - & 34.2    & 17.1    & 8.4     & 4.8    & 11.5 & 11.4 & 24.9   & 20.7    \\ \cline{2-11} 
 & S. \textit{et al.}~\cite{schwartz2019simple}   & -    & 32.1    & 16.2     & 8.7      & 5.1    & 12.1  &  11.6  & 27.6    & 21.6   \\ \cline{2-11} 
  & S. \textit{et al.}~\cite{schwartz2019simple}   & IM    & 33.8    & 16.9     & 9.1      & 5.3    & 12.7 & 11.8  & 27.7    & 22.7   \\ \cline{2-11} 
 & S. \textit{et al.}~\cite{schwartz2019simple} & MM      & 33.8    & 17.6    & 9.9     & 5.9      & 12.9 &  13.5 & 28.5      & 25.6   \\ \cline{2-11} 
  & Ours    & IM   & \textbf{37.9} & \textbf{21.6} & 12.5   & 7.6    & \textbf{15.2} & \textbf{18.5} & 31.1          & 38.1          \\ \cline{2-11} 
 & Ours    & MM      & 37.5          & 21.5          & \textbf{12.9} & \textbf{8.2} & \textbf{15.2} & 17.9 & \textbf{31.2} & \textbf{39.3} \\ \hline \hline
 
\multirow{2}{*}{\begin{tabular}[c]{@{}c@{}}Q-BOT\\ Basic baselines\end{tabular}}                     
& Ours    & IM    & 32.0     & 15.7    & 8.1     & 4.7      & 11.6 &  11.1   & 26.4    & 18.3      \\ \cline{2-11} 
 & Ours    & MM     & 33.2    & 16.4     & 8.6      & 5.0     & 12.5 & 11.5  & 27.1      & 20.2     \\ \hline
\multirow{2}{*}{\begin{tabular}[c]{@{}c@{}}Q-BOT\\ Strong baselines \end{tabular}}                    
& \begin{tabular}[c]{@{}c@{}}Ours\\(full GT HIS)\end{tabular}   & IM  & 33.3   & 17.0     & 9.1           & 5.4     & 12.6 &  11.7 & 27.3    & 21.3      \\ \cline{2-11} 
  & \begin{tabular}[c]{@{}c@{}}Ours\\(full GT HIS)\end{tabular}   & MM      & 32.7          & 17.2          & \textbf{9.7}           & \textbf{6.0}          & 12.6   &   \textbf{13.6}    & \textbf{27.9}          & \textbf{26.3}       \\ \hline

\multirow{3}{*}{\begin{tabular}[c]{@{}c@{}}Q-BOT\\ Cooperative\end{tabular}}  
&\begin{tabular}[c]{@{}c@{}}Ours\\ (pre-trained)\end{tabular} & MM     & 33.3 & 17.0          & 9.2         & 5.4         & \textbf{12.9} & 11.4 & 27.4         & 21.5         \\ \cline{2-11} 
  & Ours QA-C  & IM      & 33.3 & 16.9 & 9.1 & 5.3 & 12.7& 11.6 & 27.7 & 22.7 \\  \cline{2-11} 
&Ours QA-C & MM      &  \textbf{33.3} &   \textbf{17.3}   & 9.5     & 5.5       & 12.8 & 12.4 & \textbf{27.9}         & 23.1   \\  
\hline
\end{tabular}}
\end{center}
\label{tab:1}
\end{table}

\noindent \textbf{Description Ability for \emph{A-BOT}.} The first group of experiments focuses on the video description ability of \emph{A-BOT}, which is used as a comparison and performance reference for \emph{Q-BOT}. Under our task setup, \emph{A-BOT} has full access to all the information while \emph{Q-BOT} only sees two static images. Intuitively, \emph{A-BOT} should have better performance than \emph{Q-BOT}. We also compare the performance of our \emph{A-BOT} with the other two recent \emph{A-BOT} baselines from~\cite{hori2019end,schwartz2019simple}. The models proposed in~\cite{hori2019end,schwartz2019simple} are initially designed for question answering, but they also take all modalities of data as input, therefore, we modify the generators to generate video descriptions after 10 rounds of question-answer interactions. We observe that our \emph{A-BOT} largely outperforms the baseline models, mainly with the help of separate caption encoder and the attention operation on the dialog history, while the previous works directly incorporate the video captions into the dialog history.
Different from~\cite{schwartz2019simple} where the usage of attention for dialog history does not yield improvements for the classic question answering task, we find it is helpful in improving the performance for our video description task either by MM or IM attention method.

\begin{figure}[t]
    \centering
    \includegraphics[width=0.95\textwidth]{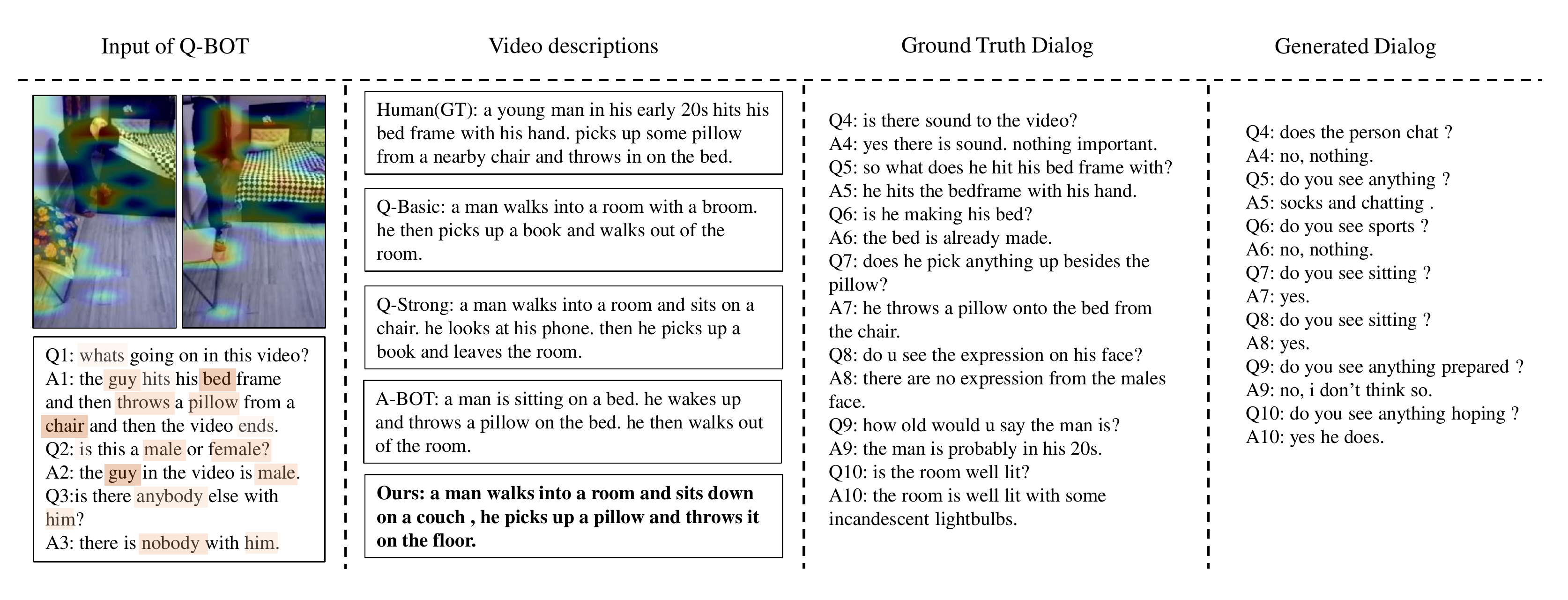}
    \caption{Example of qualitative results. Our \emph{Q-BOT} is able to describe more details about the unseen video. More qualitative results in the supplementary material.}
    \label{fig:qualitative}
\end{figure}

\noindent \textbf{Description Ability for \emph{Q-BOT}.} The second group of experiments focuses on the video description ability of \emph{Q-BOT}. The basic baseline for \emph{Q-BOT} is obtained under the standard test setting without cooperative learning, in other words, \emph{Q-BOT} does not involve question-answer interactions with \emph{A-BOT}. The comparisons between the basic baselines and \emph{A-BOT} reveal the actual knowledge gap between the two agents.
The strong \emph{Q-BOT} baselines are obtained by providing \emph{Q-BOT} with the full ground truth dialog history that contains 10 rounds of question-answer pairs. We also compare our cooperative learning method with the pre-trained methods mentioned in Sec.~\ref{subsec:learning}. The experimental results are consistent with our expectations. Although outperforming the basic baseline, the pre-trained learning method can hardly surpass two strong baselines. 
As for comparisons, our \emph{Q-BOT} from the proposed QA-Cooperative network with the cooperative learning method is able to achieve more promising performance. It outperforms the strong baseline with IM attention setting, and achieves comparable performance with the strong baseline under MM attention setting.
The experimental results demonstrate that cooperative learning helps \emph{Q-BOT} to extract even more useful information to better describe the unseen video. Moreover, we find the MM attention for the dialog history is very helpful in improving the performance of the strong baseline settings.

Examples of qualitative results are shown in Fig.~\ref{fig:qualitative}. We observe that our \emph{Q-BOT} with cooperative learning is able to describe the unseen videos with more concrete details, achieving comparable performance with the strong baseline. More qualitative results and analysis can found in the supplementary material.

\subsection{Ablation Studies}

More experimental results on ablation studies are presented in Table 3. All the performance reported is achieved by \emph{Q-BOT} under the standard test setting.

\noindent \textbf{Full Update vs. Partial Update.} In the proposed cooperative learning, the newly generated pair $r_{p,i}$ is fused with the existing dialog history $r_{H,i-1}$ before the attention module as shown in Fig.~\ref{fig:fusion}. With the updated history embedding, other outputs from the attention modules also become different compared to the previous dialog round. We compare the performance between the case of full update and partial update. In the full update setting, all the outputs from the attention modules are updated, while the partial update setting only updates the attended dialog history. The full update setting is more competitive than the partial one.

\noindent \textbf{Initial States.} In our proposed QA-Cooperative network, the initial states of the question decoder and the answer decoder are obtained from the visual LSTM and audio-visual LSTM, respectively. These two LSTM-nets summarize the audio and visual information for two agents. We observe from Table 3 that the initial states help achieve better performance under the full update setting, but have less impact on the partial update setting. It is reasonable because the full update setting mainly changes the attended audio and visual information, which is later used as the initial states. 

\begin{table}[t]
\begin{center}
\caption{Quantitative experimental results on ablation studies. All the results are obtained by \emph{Q-BOT} under the standard task setup as explained in Fig.~\ref{fig:1}.
\textit{init.} means the initial states provided by the visual and audio-visual LSTM for the question and answer decoder. \textit{w/o his for A} means we remove the attended history embedding from the input for the answer decoder.}
\scalebox{0.88}{
\begin{tabular}{|l|c|c|c|c|c|c|c|c|}
\hline
Settings           & BLEU1         & BLEU2         & BLEU3         & BLEU4        & METEOR    & SPICE    & ROUGE\_L      & CIDEr         \\ \hline
parital           & 33.2          & 16.9          & 9.2          & 5.2          & 12.8    &   12.1   & 27.6          & 21.6          \\ \hline
full w/o init.    & 33.1          &  16.3   & 8.6          & 4.9          & 12.2   &    11.6   & 27.1          & 20.3          \\ \hline
partial w/o init. & 32.7          & 16.2          & 8.8           & 5.1          & 12.1    &   11.8   & 27.4          & 20.3          \\ \hline
w/o caption       & 31.5          & 15.3          & 7.9          & 4.6          & 12.7   &   11.1    & 26.3          & 19.5          \\ \hline
w/o audio         & 33.2          & 17.2          & 9.4          & 5.4          & 12.8   &    12.2   & 27.8          & 22.3          \\ \hline
w/o his for A     & 32.5          & 16.9          & 9.3          & 5.4          & 12.1  &    12.2    & 27.1          & 23.0          \\ \hline
shuffled QA order     &      32.0     &    15.7      &     8.3      &    4.8       &  11.7 &    11.1    &     26.3      &      18.9     \\ \hline
proposed QA-C     & \textbf{33.3} & \textbf{17.3} & \textbf{9.5} & \textbf{5.5} & \textbf{12.8}& \textbf{12.4} & \textbf{27.9} & \textbf{23.1} \\ \hline
\end{tabular}}
\end{center}
\label{tab:ablation}
\end{table}

\noindent \textbf{Modality Input.} Our two agents, especially \emph{A-BOT} take multiple modalities of information as input. We test the settings when the caption and the audio information are removed from the input of \emph{A-BOT}. 
Experimental results show that the caption is of vital importance in achieving good performance for video description. Audio information also contributes to the better final performance.
\textit{w/o his for A} in Table 2 stands for the setting when the history embedding $a_{H,i-1}$ is removed from the input of answer decoder for \emph{A-BOT}. It explains the fact that \emph{A-BOT} does not heavily rely on the dialog history to provide answers to \emph{Q-BOT} since it has already watched the entire video.

\noindent \textbf{Order of Question-Answer Pairs.} We also investigate the influence of the order of the question-answer pairs in the input. Similar to~\cite{alamri2019audio}, the order of question-answer pairs is a significant factor for better final performance.

\section{Conclusions}

In summary, in this paper we propose a novel video description task via two multi-modal dialog agents, \emph{Q-BOT} and \emph{A-BOT}. We establish a new task setting for AI systems to accomplish a multi-modal task without direct access to the original visual or audio information. We further propose a QA-Cooperative network and a cooperative learning method with a dynamic dialog history update mechanism. Extensive experiments prove that \emph{Q-BOT} is able to achieve very promising performance via our proposed network and learning method.

\section*{Acknowledgements} This research was partially supported by NSF NeTS-1909185. This article solely reflects the opinions and conclusions of its authors and not the funding agents.

\clearpage
%
%
\bibliographystyle{splncs04}
\bibliography{eccv2020submissionCR}
\end{document}